\documentclass[runningheads]{llncs}

\usepackage{eccv}

\usepackage{eccvabbrv}

\usepackage{graphicx}
\usepackage{booktabs}

\usepackage[accsupp]{axessibility}  

\usepackage{hyperref}

\usepackage{orcidlink}

\usepackage{multirow} 
\usepackage{soul}
\usepackage{color}
\definecolor{lightblue}{rgb}{.8,.95,1}
\sethlcolor{lightblue}

\definecolor{red}{rgb}{1,0,0}
\definecolor{blue}{rgb}{0,0,1}
\definecolor{green}{rgb}{0.13,0.53,0.10}

\begin{document}

\title{Learning Neural Volumetric Pose Features for Camera Localization} 


\author{
    Jingyu Lin\inst{1\star}$^{\dagger}$ \and
    Jiaqi Gu\inst{2}$^{\dagger}$ \and
    Bojian Wu\inst{3} \and
    Lubin Fan\inst{2}$^{\ddagger}$ \and
    Renjie Chen\inst{1}$^{\ddagger}$ \and \\
    Ligang Liu\inst{1} \and
    Jieping Ye\inst{2}
}

\def\thefootnote{${\star}$}\footnotetext{This work was done when Jingyu Lin was an intern at Alibaba Cloud.}\def\thefootnote{\arabic{footnote}}
\def\thefootnote{${\dagger}$}\footnotetext{Equal contributions.}\def\thefootnote{\arabic{footnote}}
\def\thefootnote{${\ddagger}$}\footnotetext{Corresponding authors.}\def\thefootnote{\arabic{footnote}}

\institute{
\textsuperscript{1}University of Science and Technology of China
\textsuperscript{2}Alibaba Cloud
\textsuperscript{3}Zhejiang University
\email{sa1022@mail.ustc.edu.cn, gujiaqi.gjq@alibaba-inc.com, ustcbjwu@gmail.com, lubin.flb@alibaba-inc.com, \{renjiec, lgliu\}@ustc.edu.cn, yejieping.ye@alibaba-inc.com} \\
\url{https://gujiaqivadin.github.io/posemap/}
}

\authorrunning{J.~Lin et al.}

\maketitle

\begin{abstract}

We introduce a novel neural volumetric pose feature, termed PoseMap, designed to enhance camera localization by encapsulating the information between images and the associated camera poses. Our framework leverages an Absolute Pose Regression (APR) architecture, together with an augmented NeRF module. This integration not only facilitates the generation of novel views to enrich the training dataset but also enables the learning of effective pose features. Additionally, we extend our architecture for self-supervised online alignment, allowing our method to be used and fine-tuned for unlabelled images within a unified framework. Experiments demonstrate that our method achieves 14.28\% and 20.51\% performance gain on average in indoor and outdoor benchmark scenes, outperforming existing APR methods with state-of-the-art accuracy.
\end{abstract}    
\section{Introduction}
\label{sec:intro}

Image-based camera localization is a key task in both academic and industrial communities, which is a core component for various applications, such as reconstruction~\cite{furukawa2015multi}, navigation~\cite{biswas2012depth}, and so on. With a reference world coordinate system, the goal of camera localization is to compute the absolute camera pose, such as position and orientation, from the captured image.

Recently, Absolute Pose Regression (APR) methods have attracted more attention. Compared with traditional structure-based methods, such as Structure-from-Motion (SfM), its efficiency and better understanding of ambiguous images with textureless or repetitive patterns are more intriguing. The problem is formulated as training a regressor in a supervised manner, by taking a set of images and the associated poses estimated usually by SfM~\cite{schoenberger2016sfm}, as inputs for network training. Subsequently, the trained networks can directly regress the camera pose for the query image~\cite{chen2022dfnet}. Due to its friendly storage and computational efficiency, the APR methods are promising research directions for camera localization.

As discussed in~\cite{sattler2019understanding}, APR resembles pose approximation through image retrieval, implying that the performance is proportional to the scale of the observed scene images. With the advent of neural rendering techniques which can synthesize images from any novel viewpoints efficiently, Neural Radiance Fields (NeRF)~\cite{mildenhall2020nerf} has been incorporated into camera localization frameworks to augment both offline and online data. For instance, DFNet~\cite{chen2022dfnet} and LENS~\cite{moreau2022lens} make efforts to synthesize additional images that share the feature distribution of the training images, to narrow the domain gap between real and synthetic images. It leads to significant improvement in the localization problem.

\begin{figure}[!t]
    \centering
    \includegraphics[width=1.0\linewidth]{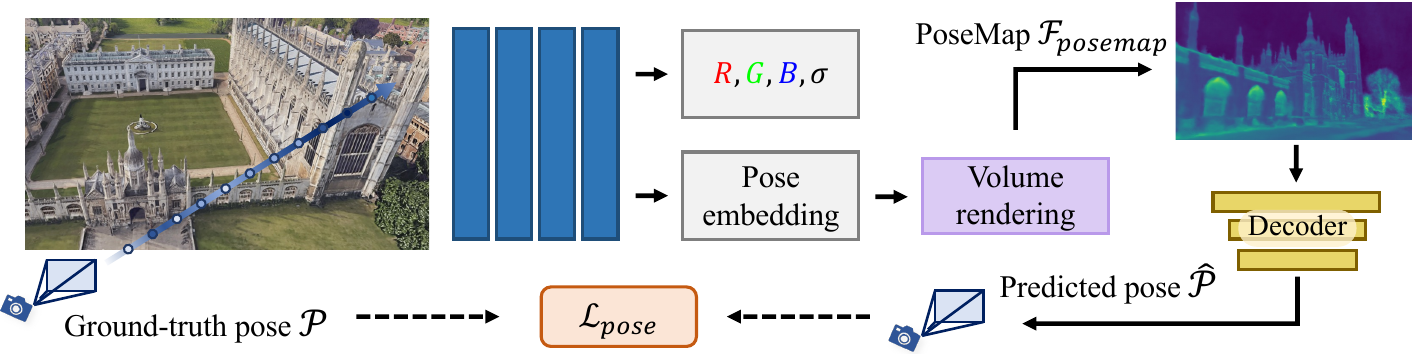}
    \caption{\textbf{The generation of PoseMap.} To capture the implicit pose characteristics, we enhance the original NeRF by introducing a unique pose embedding. Subsequently, we generate a PoseMap through volumetric rendering. We believe that the learned volumetric features integrate the implicit information of camera pose and can be used to improve the accuracy of camera localization tasks.}
    \label{fig:teaser}
\end{figure}

Our key observations are two-fold. First, the existing methods predominantly utilize the forward pass of NeRF for rendering, overlooking the potential to explore the richly encoded features within the learned volumetric field, which is an aggregation of information from images and poses, suggesting that the volumetric field of NeRF should inherently encapsulate implicit information regarding to each camera pose. Second, it is widely acknowledged that occlusions, edges, and shadows serve as explicit clues for camera localization, typically encoded as deep features within neural networks. This intuitively implies that pose estimation should be related with these feature maps.

We propose a scene-specific neural volumetric pose feature, termed PoseMap, for camera localization. As mentioned above, in order to better explore the features from the neural radiance field, by extending the original NeRF with a pose branch (NeRF-P), see Fig.~\ref{fig:teaser}, we augment the pose estimation with an additional neural feature embedding. Experimental results demonstrate that PoseMap can learn the discriminative visual and geometric features, which could distinguish query images from different viewpoints. Quantitatively, the PoseMap for APR training improves the localization accuracy by 14.28\% and 20.51\% on average over all test sets than the SOTAs. The contributions are summarized as follows.

\begin{itemize}
    \item We are the first to propose a neural volumetric pose feature, called PoseMap, for camera localization. It explores the implicit information of the image and the corresponding camera pose encoded in the neural volumetric field.
    \item We devise an APR architecture with NeRF-P, integrated with PoseMap. Both the original APR losses and pose feature loss are combined for training. We also extend our architecture for self-supervised online fine-tuning, leading to a unified framework that is also suitable for unlabelled images.
    \item Our method achieves 14.28\% and 20.51\% performance gain on average in both indoor and outdoor benchmark datasets, which outperforms the existing APR methods and sets a state-of-the-art accuracy.
\end{itemize}

\section{Related Work}
\label{sec:relatedwork}

Camera localization typically consists of two distinct stages: in-sample data acquisition and out-of-sample testing as in Table~\ref{table:summary_camloc}. The initial stage involves capturing a set of in-sample images from an unknown scene and determining the corresponding camera poses. To achieve this, the structure-based methods~\cite{schoenberger2016sfm, mur2015orb} utilize 2D keypoints matching to obtain camera poses while concurrently constructing a 3D structure. Alternatively, NeRF or 3D Gaussian Splatting based approaches~\cite{Lin2021BARFBN, Fu2023COLMAPFree3G} encapsulate and extract the pose information within their implicit models. During the out-of-sample testing phase, assuming that the poses of in-sample images have been precisely established, it is crucial to apply pose regression methods to unseen out-of-sample images, which are generally categorized into two main types. The Relative Pose Regression (RPR)-based methods~\cite{svarm2016city, toft2018semantic, sattler2016efficient, liu2017efficient, taira2018inloc} usually deduce the pose of an out-of-sample image in relation to the known poses of in-sample images, and the Absolute Pose Regression (APR)-based methods~\cite{kendall2015posenet, walch2017image, kendall2017geometric, valada2018deep, chen2022dfnet} seek to directly estimate the absolute pose within the global coordinate framework. After yielding the reliable pose prior, further refinement processes~\cite{Chen2023RefinementFA, Moreau2023CROSSFIRECR} can be invoked to iteratively enhance the camera pose estimation by leveraging pixel-wise features.

\subsection{Absolute Pose Regression (APR)}
\label{subsec:related_apr}

The Absolute Pose Regression (APR) methods, can directly predict camera pose using convolutional neural networks. These end-to-end methods exhibit notable efficiency at query time and demonstrate promising accuracy. Kendall~\etal~\cite{kendall2015posenet} proposed PoseNet, pioneering Absolute Pose Regression using a pre-trained GoogLeNet and introduced Cambridge Landmarks dataset for the task. Other variants of PoseNet, such as incorporating Bayesian methods~\cite{kendall2016modelling}, LSTM~\cite{walch2017image} and projection loss~\cite{kendall2017geometric}, have significantly enhanced the original framework's performance, since these modifications address uncertainties, capture temporal dependencies and refine pose estimation accuracy.

Confronted with limited training data, researchers have pursued two main paradigms for the improvements of APR. Sequential-based methods~\cite{valada2018deep, radwan2018vlocnet++} leverage multi-frame images to learn both absolute and relative poses by constraining the input data with temporal hints. On the other hand, rendering-based methods propose robust data augmentation to generate more training examples. For example, VGGRegNet~\cite{naseer2017deep} used synthetic depth as a reference to render and create additional viewpoints with varied pitch, zoom, and yaw settings.

Several methods have also taken into account how the representation of the pose can affect regression outcomes. For example, MapNet~\cite{brahmbhatt2018geometry} made efforts to convert rotation matrices into the logarithm of unit quaternions, aiming to ease the challenge associated with camera pose regression. RelocNet~\cite{balntas2018relocnet} introduced the concept of camera frustum overlaps to enhance the learning of pose features. Meanwhile, CamNet~\cite{ding2019camnet} suggested a bilateral frustum loss for challenging case sampling. Another approach, PAE~\cite{shavit2022camera} brought in camera pose auto-encoders as the pose representation, employing a teacher-student training strategy. 

\begin{table}[t]
    \begin{center}
    \caption{\textbf{Principle categories of techniques for the camera localization task.}}
    \label{table:summary_camloc}
    \scalebox{0.67}{
    \begin{tabular}{c|c|c|c}
    \toprule
    Stages & Category & Technology & Methods  \\
    \midrule
    \multirow{2}{*}{1st: In-sample Building}
    & 3D model-based & SfM, SLAM &  COLMAP~\cite{schoenberger2016sfm}, ORB-SLAM~\cite{Mur_Artal_2015}, ... \\
    & NeRF-based & Pose-free NeRF & BARF~\cite{Lin2021BARFBN}, COLMAP-Free 3DGS~\cite{Fu2023COLMAPFree3G}, ... \\
    \midrule
    \multirow{3}{*}{2nd: Out-of-sample Testing} 
    & Relative Regression & Image Retrieval, 2D-3D Matching & HLoc~\cite{sarlin2019coarse}, LazyVL~\cite{Dong2023LazyVL}, ...  \\
    & Absolute Regression & Absolute Pose Regression(APR)   & PoseNet~\cite{kendall2015posenet}, DFNet~\cite{chen2022dfnet}, ... \\
    & Iterative Refinement & Iterative Pose Optimization & NeFeS~\cite{Chen2023RefinementFA}, CROSSFIRE~\cite{Moreau2023CROSSFIRECR}, ... \\ 
    \bottomrule
    \end{tabular}}
    \end{center}
\end{table}

\subsection{NeRF for APR}
\label{subsec:relate_nerf}

Neural Radiance Fields (NeRF)~\cite{mildenhall2020nerf} apply differentiable volume rendering techniques to predict color and opacity from 3D position and 2D viewing direction. It has proven to be a valuable tool in a range of camera localization tasks with three main goals: offline data augmentation, online viewpoint synthesis, and feature extraction. Such as LENS~\cite{moreau2022lens} utilized a pretrained NeRF to densely sample virtual camera positions across the scene and built a synthetic dataset for training. DFNet~\cite{chen2022dfnet} adopted an online rendering strategy to render the estimated pose to a novel image and then compare the similarity of features between the real and rendered images. NeRF-loc~\cite{taira2018inloc} integrated implicit 3D descriptors from a generalizable NeRF, providing comprehensive 3D scene information. CROSSFIRE~\cite{Moreau2023CROSSFIRECR} introduced an implicit local descriptor representation for 2D-3D iterative feature matching. NeFeS~\cite{Chen2023RefinementFA} employed a post-processing pose refinement strategy with a reliable pose prior.

Recently, certain approaches~\cite{chitta2021neat,kundu2022panoptic,ding2023pdf} have been delving deeper into extracting knowledge from neural volumes, encompassing both geometrical and semantic information. In the domain of camera localization, NeRF possesses an appealing characteristic of aggregating information across multiple perspectives, which retains the associations between images and poses. Theoretically, the neural volumes of NeRF inherently encapsulate implicit information about each camera pose. 
Drawing inspiration from these insights, we directly extract pose features from neural volumes and generate a PoseMap to represent the pose information, demonstrating superior performance.

\subsection{Positioning of Our Work}

The discussions above have been exclusively focused on APR, as our method falls under this category as in Table~\ref{table:summary_camloc}. In terms of alternative techniques related to camera localization, we have classified them into several principal categories. 
We emphasize that APR methods can deduce camera poses from unseen images using a single inference pass of a network, obviating the need for 2D-2D/3D matching with pre-built models or pose priors.
\section{Method}
\label{sec:method}

\begin{figure*}[t]
    \centering
    \includegraphics[width=1.0\linewidth]{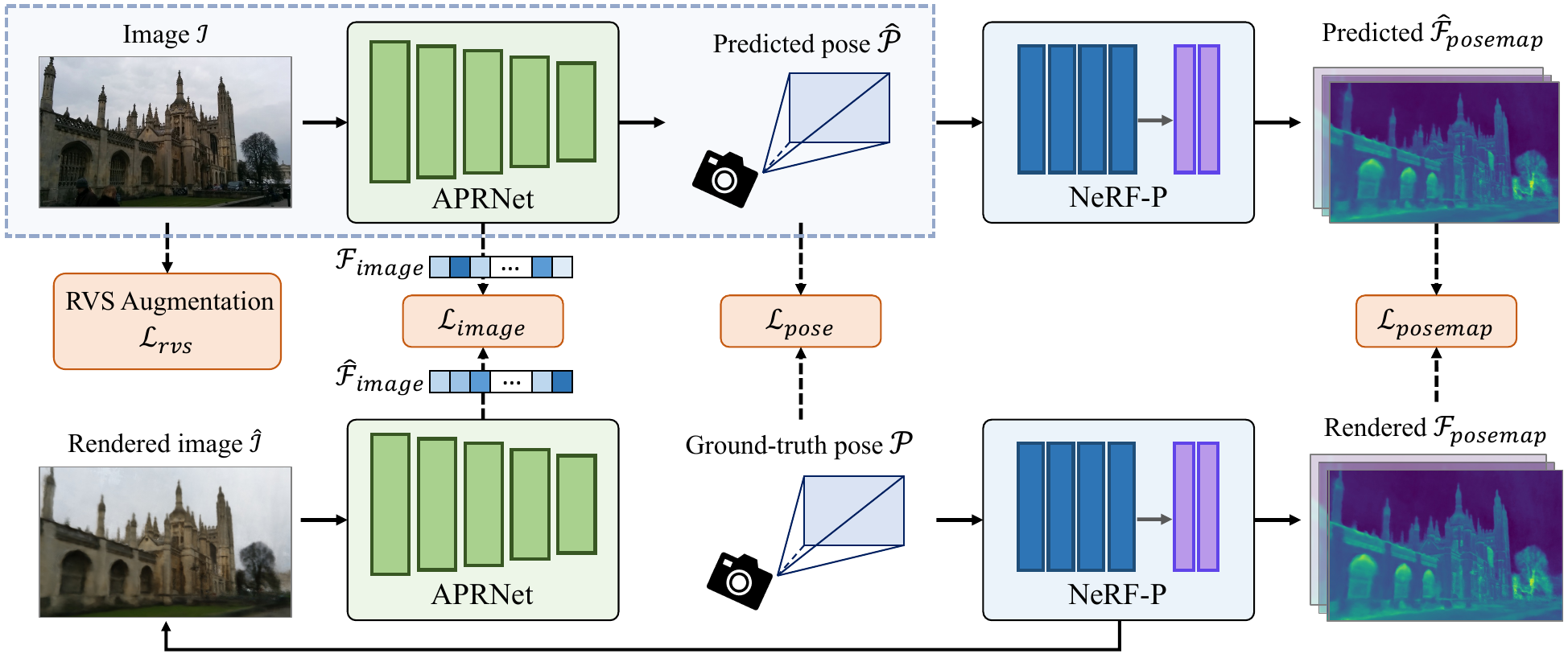}
    \caption{\textbf{Overview of the camera localization pipeline with PoseMap}. The training stage of our pipeline, including two main modules: APRNet for camera pose regression and extracting image features and NeRF-P for view synthesis and extracting pose features. The inference stage of our pipeline with a simple APRNet for fast inference is highlighted in blue dotted box.}
    \label{fig:overview}
\end{figure*}

\textbf{Overview.}
Given a set of images and the associated camera poses $\{(I, p)\}$, our goal is to train a neural network that takes a query image $I^*$ as input and directly outputs its corresponding camera pose $p^*$. Here the camera pose is represented as a $3\times4$ matrix which is a combination of translation and rotation with regard to a reference coordinate system. The whole pipeline is demonstrated in Fig.~\ref{fig:overview}. It contains two main entangled modules: APRNet and NeRF-P. Concretely, with an input image $I$, APRNet leverages separate branches to extract image features $\mathcal{F}_{image}(I)$ and estimates the camera pose $\hat{p}$. With the given ground-truth pose, NeRF-P subsequently renders a synthetic image $\hat{I}$, which will be forwarded to the feature extraction branch of APRNet and obtain $\mathcal{F}_{image}(\hat{I})$. 
On the other hand, we propose a novel implicit pose features $\mathcal{F}_{pose}(\hat{p})$ called \textit{PoseMap}, by enhancing the volumetric rendering module with an extra pose feature branch on original NeRF architecture, which will be further used for pose prediction. The key idea of this design choice is that NeRF itself is an abstraction of visual and geometric information. We propose an autoencoder-style pose branch to leverage NeRF to aggregate global attributes from samples of light rays on PoseMap and a pose decoder of a series of MLP decoders for the distillation of implicit pose features. This novel combination allows for a more precise and detailed representation of the camera pose. In all, APRNet is optimized, with the supervision of the ground-truth PoseMap $\mathcal{F}_{pose}(p)$,  and by minimizing the discrepancies of image features. This section is organized as follows: the camera localization pipeline and its core modules are discussed in Sec.~\ref{subsec:pipeline}, the neural volumetric pose feature PoseMap is detailed in Sec.~\ref{subsec:nerfp}, followed by a self-supervised scheme for online feature alignment using unlabelled images with the guidance of PoseMap in Sec.~\ref{subsec:self_supervised}.

\subsection{Camera Localization with PoseMap}
\label{subsec:pipeline}

\textbf{Architecture.}
Our proposed camera localization architecture (Fig.~\ref{fig:overview}) consists of two main modules: \textit{APRNet} and \textit{NeRF-P}, as introduced in the overview.

\textit{APRNet} comprises two sub-networks: a pose estimator and an image feature extractor. The pose estimator predicts the camera pose $\hat{p}$ for an input image $I$. Meanwhile, the feature extractor generates deep features $\mathcal{F}_{image}(I)$ of $I$. The architecture of the APRNet is shown in Fig.~\ref{fig:overview}. Similar to~\cite{chen2022dfnet}, the final image features are fused by feature maps from each convolutional block in the pose estimator. During the training stage, the pose estimator network is refined by aligning image features from the feature extractors. In the inference stage, the pose estimator is applied directly.

\textit{NeRF-P} includes two distinct branches: the rendering branch and the volumetric pose extraction branch for generating PoseMap, which is the key component of our design. Unlike CROSSFIRE~\cite{Moreau2023CROSSFIRECR} and NeFeS~\cite{Chen2023RefinementFA} that extend NeRF to synthesize CNN image features, we focuses on establishing a direct relationship between implicit 3D information and camera poses. The details will be described in Sec.~\ref{subsec:nerfp}.

\textbf{Pipeline.}
Given a set of input images paired with the camera poses $\{(I, p)\}$, NeRF-P undergoes a two-step training process. Initially, to maintain the quality of the original NeRF, the rendering branch is trained without considering pose features. Subsequently, the learned weights of the network are frozen, and the pose branch is trained independently. Utilizing the input image $I$, APRNet estimates a camera pose $\hat{p}$, and NeRF-P then generates the PoseMap $\mathcal{F}_{pose}(\hat{p})$ from $\hat{p}$. Conversely, with the ground-truth camera pose of $I$, NeRF-P also generates a PoseMap $\mathcal{F}_{pose}(p)$ and produces a synthetic image $\hat{I}$. The discrepancies between the image features from the real image $I$ and the synthetic image $\hat{I}$, as well as the differences between pose features $\mathcal{F}_{pose}(\hat{p})$ and $\mathcal{F}_{pose}(p)$, are computed. The pose estimator of APRNet is optimized by minimizing domain gaps in both image features and pose features. To quantify the differences in image features, we apply the triplet loss from~\cite{chen2022dfnet} as the image feature loss $\mathcal{L}_{image}$. In practice, we employ $L_2$ loss to compute $\mathcal{L}_{pose}$ and adopt cosine similarity loss for $\mathcal{L}_{posemap}$.

Considering the scalability of performance related to the number of observed images, we apply the NeRF-P module to synthesize more unseen views from randomly perturbed poses, named random view synthesis (RVS). Given a camera pose $p$, a perturbed pose $p_{rvs}$ is generated with a random translation and rotation. 
Given a synthetic image $I_{rvs}$ with the perturbed pose $p_{rvs}$, the APRNet estimates its camera pose $\hat{p}_{rvs}$, followed by the NeRF-P to extract its pose feature map $\mathcal{F}_{rvs}$. 
The Same supervising strategies then are applied with the loss function defined as
\begin{equation}
\mathcal{L}_{rvs} = \| \hat{p}_{rvs} - p_{rvs} \|_2 + \| \hat{\mathcal{F}}_{rvs} - \mathcal{F}_{rvs} \|_2 .
\end{equation}

\textbf{Loss Function.}
To summarize, the total loss for optimizing the pose estimator in the APRNet is: 
\begin{equation}
\begin{aligned}
    \mathcal{L}_{total} &= \mathcal{L}_{pose} + \mathcal{L}_{image} + \mathcal{L}_{posemap} + \mathcal{L}_{rvs}.
\end{aligned}
\label{eq:loss_total}
\end{equation}

\subsection{Neural Volumetric Features of PoseMap}
\label{subsec:nerfp}

Here, we introduce the detailed design of PoseMap, which is generated from the pose branch of the augmented NeRF, aka NeRF-P.

Given an image $I$, its appearance depends on the geometric and visual features of the scene and the camera settings, especially the camera pose. Intuitively, the visual clues, such as edges, shadows, and semantics, are proven to be crucial to camera localization. Thus, the existing methods leverage convolutional neural networks to extract deep features and subsequently predict the camera poses. However, most of the current methods are data-driven, which often struggle to deal with the data with sparse distribution because of a lack of global observation and any prior information on the whole scene.

Inspired by NeRF, which is a natural fusion of both images and poses, the volumetric representation of the scene can provide strong priors for the downstream applications, and avoid the aforementioned problems. Along with the neural rendering pipeline, we seek to explore the neural radiance field and separately abstract pose features from the volumetric feature space. Concretely, we learn a $k$-dimensional pose features vector for each sampled point on the camera ray and also apply a neural rendering approach to calculate the integrated features for each pixel, which is defined as PoseMap in our case.
We sample $N$ points along the ray and approximate the pose feature vector $\hat{F}$ by
\begin{equation}
    \hat{F} = \sum_{i=1}^N w_i f_i,
\end{equation}
\begin{equation}
    \text{with}~w_i = T_i \left( 1 - \exp(-\sigma_i \delta_i) \right), T_i = \exp\left( -\sum_{j=1}^{i-1} \sigma_j \delta_j \right),
\end{equation}
where $w_i$ is the weight of rendering, $\sigma_i$ and $f_i$ denote the density and pose feature vector at each sampled point $i$ on the ray, and $\delta_j$ is the distance between adjacent samples. Fig.~\ref{fig:teaser} illustrates an example of PoseMap.

\begin{figure}[t]
    \centering
    \includegraphics[width=1.0\linewidth]{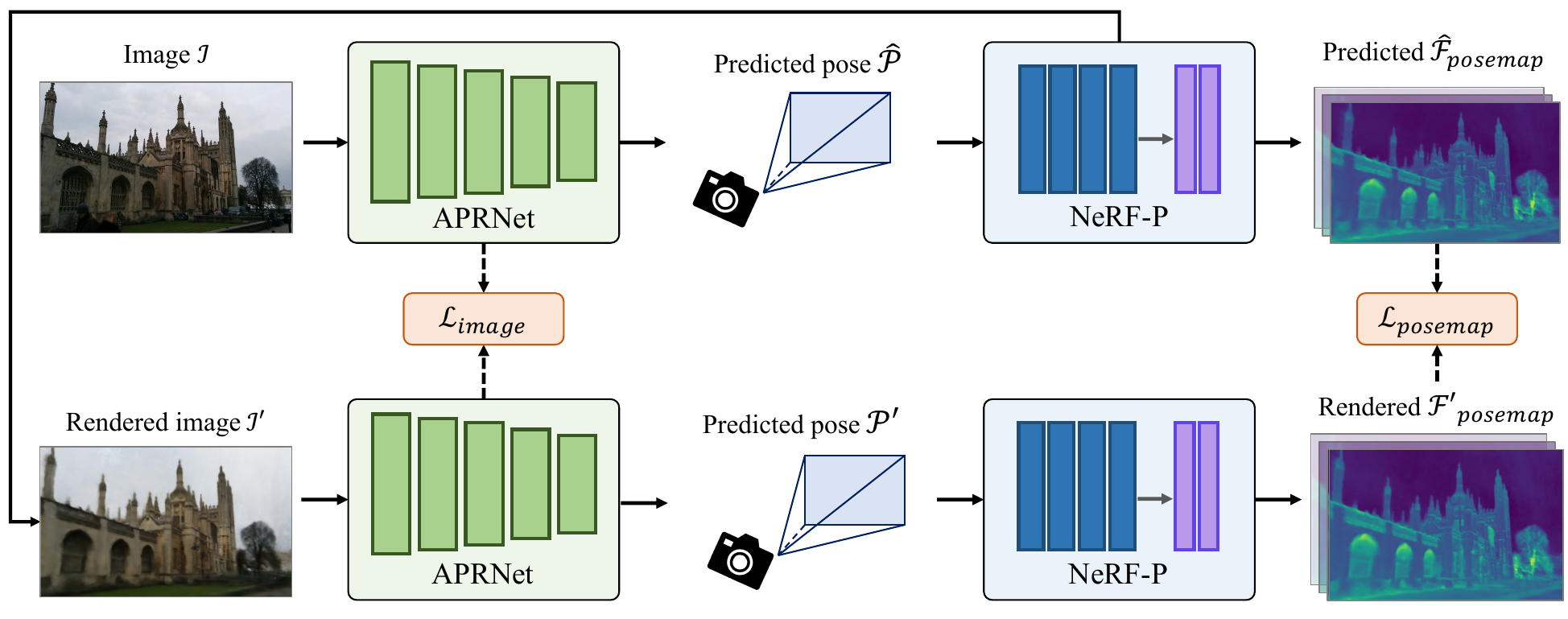}
    \caption{\textbf{Self-supervised online feature alignment scheme.} We keep the $\mathcal{L}_{image}$ and $\mathcal{L}_{posemap}$ in the self-supervised pipeline. This scheme is suitable for any unlabelled images or images from the internet without matching with the 3D SfM model. }
    \label{fig:self-supervised}
\end{figure}

\subsection{Self-supervised Feature Alignment with PoseMap}
\label{subsec:self_supervised}

As previously noted, the accuracy of the APR-based method tends to correlate with the size of the observed scene images. 
We not only randomly synthesize labelled images but also incorporate unlabelled images with unknown camera poses to further enhance the training dataset,
we propose a self-supervised scheme aimed at refining the pre-trained APRNet with PoseMap(refer to Fig.~\ref{fig:self-supervised}). When presented with an unlabelled image $I$, we utilize the pre-trained APRNet to estimate its camera pose, denoted as $\hat{p}$. Subsequently, the PoseMap $\mathcal{F}_{pose}(\hat{p})$ is generated from NeRF-P using $\hat{p}$, and a new unlabelled image $I'$ is rendered. We follow the same procedure for image $I'$ as we do for $I$, with the exception of omitting random view synthesis. The refinement of APRNet involves minimizing gaps in both image features and pose features. The loss for APRNet online feature alignment is
\begin{equation}
\begin{aligned}
    \mathcal{L}_{align} &= \mathcal{L}_{image} + \mathcal{L}_{posemap}.
\end{aligned}
\label{eq:loss_self}
\end{equation}

\section{Experiments}
\label{sec:experiments}

In this section, we evaluate our camera localization method combined with pose features on standard benchmarks and perform comprehensive studies to demonstrate the effectiveness of our design.

\subsection{Datasets and Implementation Details}
\label{subsec:exp_setup}

\textbf{Datasets.} 
We evaluate our method on two visual localization datasets covering both indoor and outdoor scenes:
\begin{itemize}
    \item 7-scenes~\cite{shotton2013scene}: it consists of 7 small-scale (about $1\sim18m^2$) static indoor scenes and up-to 7000 training images that are captured by a Kinect RGB-D sensor. The test sequences have completely different paths from train sequences which make it to be a difficult dataset for camera estimation tasks.
    \item Cambridge Landmarks~\cite{kendall2015posenet}: it contains 4 large-scale (about $900 \sim 5500m^2$) dynamic outdoor scenes with images and large SfM reconstructions, including all the recovered camera poses. The training sequences contain $200\sim1500$ samples, and testing sets are captured from different sequences.
\end{itemize}
For all studies, we use the ground-truth image-pose pairs for training. For each scene, camera poses are obtained from SfM or SLAM and later re-aligned and re-centered in $SE(3)$ to relieve the scale and distribution bias. In the following experiments, we report the median translation (m) and rotation ($^\circ$) errors for evaluations. 

\textbf{NeRF-P settings.} 
We implement NeRF-P based on the source code from~\cite{chen2022dfnet} and use the default parameters for fore-/background separation. An MLP encoder after the static branch is implemented to recover a pose embedding. Then the PoseMap is rendered as the size of $H \times W \times C$, where $C$ is 256 according to experiments. The PoseMap is downsampled by a factor of 16 with the $H \times W$ dimensions flattened, followed by 4 layers of MLPs $(1536, 256, 128, 12)$ to yield the predicted pose. Practically, we first train the NeRF-P without the pose branch, then freeze the main weights and only train the pose branch and decoder.

\textbf{Pipeline settings.} 
We train the network on each dataset from scratch. The APRNet adapts a pre-trained VGG-16 on ImageNet as the backbone and applies the Adam optimizer with a learning rate of $1 \times 10^{-4}$ for training. The image feature embedding is extracted following the procedure as~\cite{chen2022dfnet}. To synthesize random views, we add random perturbations (i.e., $\delta_t=[0.2,0.2,0.2]m, \delta_r=[10,10,10]^\circ$ for indoor scenes and $\delta_t=[3.0,3.0,3.0]m, \delta_r=[7.5,7.5,7.5]^\circ$ for outdoor scenes) on the training poses. For fair comparison to other methods, we use 20\%$/$50\% of the original data for indoor$/$outdoor scenes training, respectively.

\textbf{Self-supervised settings.}
At the online feature alignment stage, we load the pre-trained weights of APRNet and refine it with images without ground-truth camera poses. Following the setting in~\cite{ brahmbhatt2018geometry, chen2021direct, chen2022dfnet}, we use 10\% (indoor) and 50\% of images (outdoor) without ground-truth poses from the validation dataset, respectively. The batch size is 1 and the learning rate is $1 \times 10^{-5}$. We name the pipeline before and after self-supervised alignment with unlabelled data as \textbf{PMNet} and \textbf{PMNet$_{ud}$}, respectively.

\textbf{Other details}.
We utilize a multi-stage training schedule, starting from NeRF-P, followed by APRNet using labeled images, and subsequently incorporating unlabeled images. Additionally, we generate novel views using an online RVS strategy from the synthesized poses within the batch samples, along with real images. The inference time for each image ($480\times854$) is 6.27ms on average. All experiments are conducted on NVIDIA RTX 4090 GPU with PyTorch. 

\begin{table*}[t!]
    \begin{center}
    \caption{\textbf{Single-frame APR results on 7-scenes dataset.} We compared our pipeline with prior single-frame APR methods in median translation (m) and rotation error ($^\circ$). For better visualization, the best results are highlighted in \hl{\textbf{bold}}.}
    \label{table:7scenes}
    
    \scalebox{0.6}{
    \begin{tabular}{c|c|c|c|c|c|c|c|c|c}
    \toprule
    Category & Methods & Chess & Fire & Heads & Office & Pumpkin & Kitchen & Stairs & Average  \\
    \cmidrule{1-10}
    \multirow{16}{*}{Single-frame APR} 
    & PoseNet(PN)~\cite{kendall2015posenet}            & 0.32/8.12 & 0.47/14.4 & 0.29/12.0 & 0.48/7.68 & 0.47/8.42 & 0.59/8.64 & 0.47/13.8 & 0.44/10.4 \\
    & BayesianPN~\cite{kendall2016modelling}           & 0.37/7.24 & 0.43/13.7 & 0.31/12.0 & 0.48/8.04 & 0.61/7.08 & 0.58/7.54 & 0.48/13.1 & 0.47/9.81 \\
    & PN Learn $\sigma^2$~\cite{kendall2017geometric}  & 0.14/4.50 & 0.27/11.8 & 0.18/12.1 & 0.20/5.77 & 0.25/4.82 & 0.24/5.52 & 0.37/10.6 & 0.24/7.87 \\
    & geo. PN~\cite{kendall2017geometric}              & 0.13/4.48 & 0.27/11.3 & 0.17/13.0 & 0.19/5.55 & 0.26/4.75 & 0.23/5.35 & 0.35/12.4 & 0.23/8.12 \\
    & LSTM PN~\cite{walch2017image}                    & 0.24/5.77 & 0.34/11.9 & 0.21/13.7 & 0.30/8.08 & 0.33/7.00 & 0.37/8.83 & 0.40/13.7 & 0.31/9.85 \\
    & Hourglass PN~\cite{melekhov2017image}            & 0.15/6.17 & 0.27/10.8 & 0.19/11.6 & 0.21/8.48 & 0.25/7.00 & 0.27/10.2 & 0.29/12.5 & 0.23/9.53 \\
    & BranchNet~\cite{wu2017delving}                   & 0.18/5.17 & 0.34/8.99 & 0.20/14.2 & 0.30/7.05 & 0.27/5.10 & 0.33/7.40 & 0.38/10.3 & 0.28/8.30 \\
    & GPoseNet~\cite{cai2019hybrid}                    & 0.20/7.11 & 0.38/12.3 & 0.21/13.8 & 0.28/8.83 & 0.37/6.94 & 0.35/8.15 & 0.37/12.5 & 0.31/9.95 \\
    & MapNet~\cite{brahmbhatt2018geometry}             & 0.08/3.25 & 0.27/11.7 & 0.18/13.3 & 0.17/5.15 & 0.22/4.02 & 0.23/4.93 & 0.30/12.1 & 0.21/7.77 \\
    & Direct-PN~\cite{chen2021direct}                  & 0.10/3.52 & 0.27/8.66 & 0.17/13.1 & 0.16/5.96 & 0.19/3.85 & 0.22/5.13 & 0.32/10.6 & 0.20/7.26 \\
    & TransPoseNet~\cite{shavit2021paying}             & 0.07/5.68 & 0.24/10.6 & 0.13/12.7 & 0.17/6.34 & 0.17/5.60 & 0.19/6.75 & 0.30/7.02 & 0.18/7.78 \\
    & MSPN~\cite{blanton2020extending}                 & 0.09/4.76 & 0.29/10.5 & 0.16/13.1 & 0.16/6.80 & 0.19/5.50 & 0.21/6.61 & 0.31/11.6 & 0.20/8.41 \\
    & MS-Transformer~\cite{shavit2021learning}         & 0.11/4.66 & 0.24/9.60 & 0.14/12.2 & 0.17/5.66 & 0.18/4.44 & 0.17/5.94 & 0.17/5.94 & 0.18/7.28 \\
    & PAE~\cite{shavit2022camera}                      & 0.12/4.95 & 0.24/9.31 & 0.14/12.5 & 0.19/5.79 & 0.18/4.89 & 0.18/6.19 & 0.25/8.74 & 0.19/7.48 \\
    & DFNet~\cite{chen2022dfnet}                       & 0.05/1.88 & 0.17/6.45 & \hl{\textbf{0.06}/\textbf{3.63}} & 0.08/2.48 & \hl{\textbf{0.10}/\textbf{2.78}} & 0.22/5.45 & \hl{\textbf{0.16}/\textbf{3.29}} & 0.12/3.71 \\
    & \textbf{PMNet(Ours)}                                      & \hl{\textbf{0.04}/\textbf{1.70}} & \hl{\textbf{0.10}/\textbf{4.51}} & 0.07/4.23 & \hl{\textbf{0.07}/\textbf{1.96}} & 0.14/3.33 & \hl{\textbf{0.14}/\textbf{3.36}} & \hl{\textbf{0.16}}/3.62 & \hl{\textbf{0.10}/\textbf{3.24}} \\
    \midrule
    \multirow{6}{*}{APR with Unlabelled Data.} 
    & MapNet$_{+(seq.)}$~\cite{brahmbhatt2018geometry}     & 0.10/3.17 & 0.20/9.04 & 0.13/11.1 & 0.18/5.38 & 0.19/3.92 & 0.20/5.01 & 0.30/13.4 & 0.19/7.29 \\
    & MapNet$_{+PGO(seq.)}$~\cite{brahmbhatt2018geometry}      & 0.09/3.24 & 0.20/9.29 & 0.12/8.45 & 0.19/5.42 & 0.19/3.96 & 0.20/4.94 & 0.27/10.6 & 0.18/6.55 \\
    & Direct-PN+U~\cite{chen2021direct}                        & 0.09/2.77 & 0.16/4.87 & 0.10/6.64 & 0.17/5.04 & 0.19/3.59 & 0.19/4.79 & 0.24/8.52 & 0.24/8.52 \\
    & LENS$_{+(seq.)}$~\cite{moreau2022lens}                               & \hl{\textbf{0.03}}/1.30 & 0.10/3.70 & 0.07/5.80 & 0.08/1.90 & 0.08/2.20 & 0.09/\hl{\textbf{2.20}} & 0.14/3.60 & 0.08/2.96 \\
    & DFNet$_{dm}$~\cite{chen2022dfnet}                        & 0.04/1.48 & \hl{\textbf{0.04}}/2.16 & 0.03/1.82 & 0.07/2.01 & 0.09/2.26 & 0.09/2.42 & 0.14/3.31 & 0.07/2.21 \\
    & \textbf{PMNet$_{ud}$(Ours)}                                     & \hl{\textbf{0.03}/\textbf{1.26}} & \hl{\textbf{0.04}/\textbf{1.76}} & \hl{\textbf{0.02}/\textbf{1.68}} & \hl{\textbf{0.06}/\textbf{1.69}} & \hl{\textbf{0.07}/\textbf{1.96}} & \hl{\textbf{0.08}}/2.23 & \hl{\textbf{0.11}/\textbf{2.97}} & \hl{\textbf{0.06}/\textbf{1.93}} \\
    \bottomrule
    \end{tabular}}
    \end{center}
\end{table*}

\begin{figure*}[!t]
    \centering
    \includegraphics[width=\linewidth]{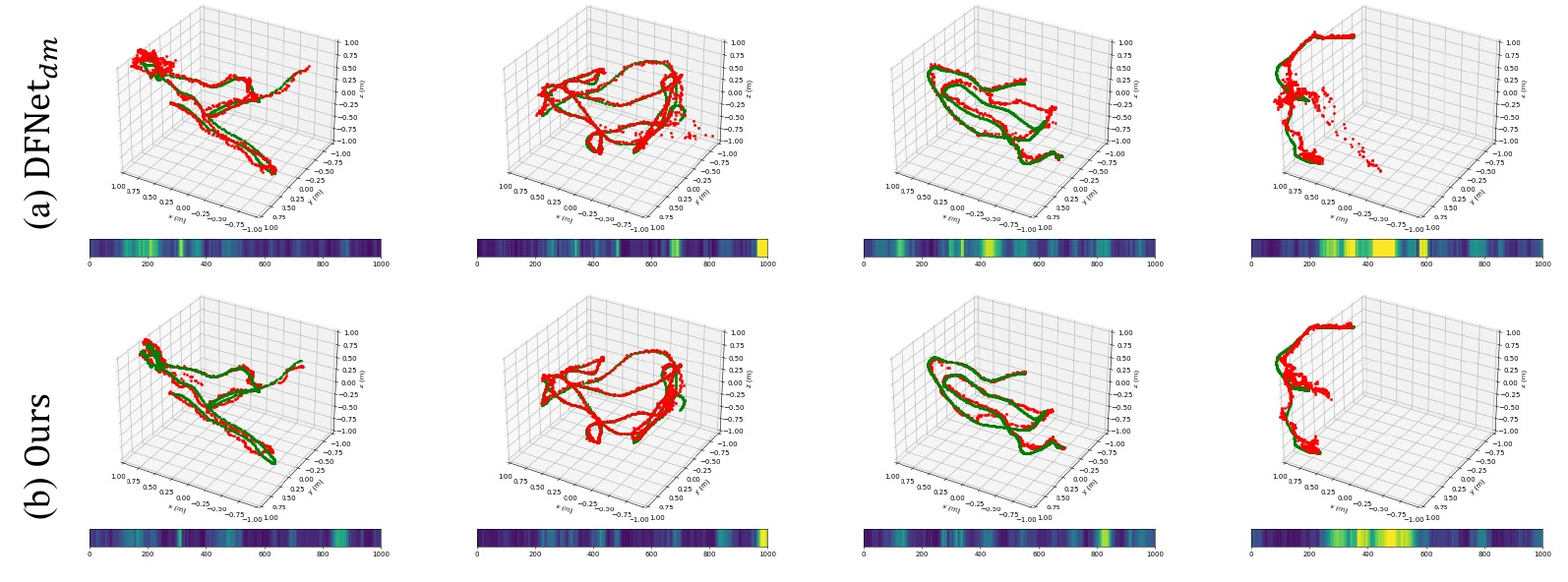}
    \caption{\textbf{Visual comparison of camera localization between DFNet$_{dm}$ (top) and our method (bottom) on 7-scenes dataset.} For each plot, we show the ground truth camera trajectory in \textbf{\textcolor{green}{green}} and the estimated trajectory in \textbf{\textcolor{red}{red}}. The color bar under each plot shows rotation errors. Yellow represents high rotation error, and blue represents low rotation error. Sequence names from left to right are: office-seq7, chess-seq3, fire-seq3 and kitchen-seq4.}
    \label{fig:track}
\end{figure*}

\subsection{Evaluation on Datasets}
\label{subsec:exp_eval}

\textbf{Evaluation on 7-scenes dataset.} 
First, we compare our methods with one-frame APR methods on 7-scenes dataset. Table~\ref{table:7scenes} shows that our method achieves the best results on all indoor scenes.
To be specific, our main pipeline, denoted as PMNet, achieves state-of-the-art performance. Moreover, PMNet$_{ud}$ finetuned in the self-supervising stage with unlabelled data further enhances the pose regression accuracy. The translation and rotation errors are reduced by 14.28\% (0.07$\rightarrow$0.06) and 12.67\%(2.21$\rightarrow$1.93) on average, respectively. 
For the Heads and Pumpkin scenes, our results are comparable to those of DFNet, though not superior. Upon reviewing the dataset, we found that these scenes are relatively small, with limited visible range and camera pose variations. Consequently, the benefits of our PoseMap may not be particularly significant.

We visualize the camera localization sequences on 7-scenes dataset in Fig.~\ref{fig:track}. It shows that our trajectories mostly coincide with the ground truth that means our method could estimate camera positions with high accuracy. Comparing with the SOTA method (i.e., DFNet), our method achieves better estimation results since our trajectories are much closer to the ground truth, and the number of high rotation error zones is less.

\begin{table}[t!]
    \begin{center}
    \caption{\textbf{Single-frame APR results on Cambridge Landmarks dataset.} We compare with APR-based methods without iterative refinement and report the median position and orientation error in m/$^\circ$.  The best results are highlighted in \hl{\textbf{bold}}.}
    \label{table:cambridge}
    \scalebox{0.75}{
    \begin{tabular}{c|c|c|c|c|c|c}
    \toprule
    Category & Methods & Kings & Hospital & Shop & Church & Average  \\
    \midrule
    \multirow{12}{*}{Single-frame APR} 
    & PoseNet(PN)~\cite{kendall2015posenet}            & 1.66/4.86 & 2.62/4.90 & 1.41/7.18 & 2.45/7.95 & 2.04/6.23 \\
    & BayesianPN~\cite{kendall2016modelling}           & 1.74/4.06 & 2.57/5.14 & 1.25/7.54 & 2.11/8.38 & 1.92/6.28 \\
    & PN Learn $\sigma^2$~\cite{kendall2017geometric}  & 0.99/1.06 & 2.17/2.94 & 1.05/3.97 & 1.49/3.43 & 1.43/2.85 \\
    & geo. PN~\cite{kendall2017geometric}              & 0.88/\hl{\textbf{1.04}} & 3.20/3.29 & 0.88/3.78 & 1.57/\hl{\textbf{3.32}} & 1.63/2.86 \\
    & LSTM PN~\cite{walch2017image}                    & 0.99/3.65 & 1.51/4.29 & 1.18/7.44 & 1.52/6.68 & 1.30/5.51 \\
    & GPoseNet~\cite{cai2019hybrid}                    & 1.61/2.29 & 2.62/3.89 & 1.14/5.73 & 2.93/6.46 & 2.08/4.59 \\
    & MapNet~\cite{brahmbhatt2018geometry}             & 1.07/1.89 & 1.94/3.91 & 1.49/4.22 & 2.00/4.53 & 1.63/3.64 \\
    & MSPN~\cite{blanton2020extending}                 & 1.73/3.65 & 2.55/4.05 & 2.02/7.49 & 2.67/6.18 & 2.47/5.34 \\
    & MS-Transformer~\cite{shavit2021learning}         & 0.83/1.47 & 1.81/2.39 & 0.86/3.07 & 1.62/3.99 & 1.28/2.73 \\
    & PAE~\cite{shavit2022camera}                      & 0.90/1.49 & 2.07/2.58 & 0.99/3.88 & 1.64/4.16 & 1.40/3.03 \\
    & DFNet~\cite{chen2022dfnet}                       & 0.73/2.37 & 2.00/2.98 & 0.67/2.21 & 1.37/4.03 & 1.19/2.90 \\
    & \textbf{PMNet(Ours)}                                      & \hl{\textbf{0.68}}/1.97 & \hl{\textbf{1.03}/\textbf{1.31}} & \hl{\textbf{0.58/2.10}} & \hl{\textbf{1.33}}/3.73 & \hl{\textbf{0.90}/\textbf{2.27}} \\
    \midrule
    \multirow{3}{*}{APR with Unlabelled Data.} 
    & LENS$_{+(seq.)}$~\cite{moreau2022lens}                       & 0.33/\hl{\textbf{0.50}}  & \hl{\textbf{0.44}}/0.90 & 0.27/1.60 & 0.53/1.60 & 0.39/1.15 \\ 
    & DFNet$_{dm}$~\cite{chen2022dfnet}                & 0.43/0.87 & 0.46/0.87 & \hl{\textbf{0.16}/\textbf{0.59}} & 0.50/1.49 & 0.39/0.96 \\
    & \textbf{PMNet$_{ud}$(Ours)}                    & \hl{\textbf{0.31}}/0.55 & \hl{\textbf{0.44}/\textbf{0.79}} & 0.17/0.86 & \hl{\textbf{0.31}/\textbf{0.96}} & \hl{\textbf{0.31}/\textbf{0.79}} \\
    \midrule
    \midrule
    \multirow{4}{*}{APR + Iterative Refinement.} 
    & DFNet~\cite{chen2022dfnet}+NeFeS$_{30}$~\cite{Chen2023RefinementFA}   & 0.37/0.64 & 0.98/1.61 & 0.17/0.60 & 0.42/1.38 & 0.48/1.06 \\
    & DFNet~\cite{chen2022dfnet}+NeFeS$_{50}$~\cite{Chen2023RefinementFA}   & 0.37/0.62 & 0.55/0.90 & 0.14/0.47 & 0.32/0.99 & 0.35/0.75 \\
    & CROSSFIRE~\cite{Moreau2023CROSSFIRECR}           & 0.47/0.70 & 0.43/0.70 & 0.20/1.20 & 0.39/1.40 & 0.37/1.00 \\
    & HR-APR~\cite{Liu2024HRAPRAF}                     & 0.36/0.58 & 0.53/0.89 & 0.13/0.51 & 0.38/1.16 & 0.35/0.78 \\
    \bottomrule
    \end{tabular}}
    \end{center}
\end{table}

\textbf{Evaluation on Cambridge landmarks dataset.} 
Then we evaluate our methods on the more challenging dataset, i.e., Cambridge Landmarks. As shown in Table~\ref{table:cambridge}, our PMNet leads to the dominant performance advantage over all scenes in single-frame APR methods. Compared to DFNet$_{dm}$~\cite{chen2022dfnet} which also uses unlabelled data for online training, our method gets performance gain by 20.51\% (0.39$\rightarrow$0.31) and 17.71\% (0.96$\rightarrow$0.79) for translations and rotations, respectively. 

\begin{table}[t!]
    \begin{center}
    \caption{\textbf{Camera localization results on TOP 30 outliers of camera poses in the testing split on Cambridge Landmarks dataset.} We report the mean translation and rotation error in m/$^\circ$. The better results are highlighted in \hl{\textbf{bold}}.}
    \label{table:ood}
    \scalebox{0.8}{
    \begin{tabular}{c|c|c|c|c|c}
    \toprule
    Methods & Kings & Hospital & Shop & Church & Average  \\
    \midrule
    DFNet$_{dm}$~\cite{chen2022dfnet}  & 5.17/3.25 & \hl{\textbf{0.55}}/1.11 & \hl{\textbf{0.34}}/1.53 & 1.72/3.42 & 1.94/2.32 \\
    PMNet$_{ud}$(Ours)               & \hl{\textbf{1.61}}/\hl{\textbf{1.65}} & 0.91/\hl{\textbf{1.01}} & 0.37/\hl{\textbf{1.48}} & \hl{\textbf{0.54}}/\hl{\textbf{1.67}} & \hl{\textbf{0.85}}/\hl{\textbf{1.45}} \\
    \bottomrule
    \end{tabular}}
    \end{center}
\end{table}

Besides the overall performance, it's also crucial to achieve high accuracy for frames with sparse distribution, such as images taken far from training views. To evaluate the robustness, we perform spatial clustering on the distribution of camera positions of the training set and calculate the Hausdorff distance from each camera pose of the testing set to the nearest cluster center, which we refer to as the outlier distance. Camera poses of these testing images have the largest outlier distance from the training set. We experimented with 30 images selected from the testing set in Kings of Cambridge Landmarks. Table~\ref{table:ood} shows that PMNet$_{ud}$ obtains lower average errors than DFNet with both position (0.85 vs. 1.94) and orientation (1.45 vs. 2.32). It demonstrates that PoseMap could reinforce the network to learn and explore the less observed views.

\begin{table}[t!]
\centering
\caption{\textbf{Ablation results on different loss settings in Kings of Cambridge Landmarks.} Table~\ref{tab:ablation_suba} shows the exploration of different pose features of PMNet. In Table~\ref{tab:ablation_subb}, we ablate different settings of both PMNet and PMNet$_{ud}$ in median translation (m) and rotation error ($^\circ$). Note that Exp. 5 and 6 of PMNet$_{ud}$ are finetuned on the pretrained model of Exp. 4 of PMNet.}
\label{tab:my_table}
\scalebox{0.7}{
    \begin{minipage}{0.44\textwidth} 
        \begin{subtable}{\linewidth}
            \centering
            \caption{The ablation of pose features.}
            \label{tab:ablation_suba}
            \begin{tabular}{c|c}
            \toprule
            Pose Feature & Performance \\
            \midrule
            None  & 1.03~/~2.94 \\
            with $\mathcal{L}_{nerfmap}$  & 0.95~/~5.76 \\
            with $\mathcal{L}_{posemap}$  & \hl{\textbf{0.68~/~1.97}} \\
            \bottomrule
            \end{tabular}
        \end{subtable}%
    \end{minipage}
    \begin{minipage}{0.52\textwidth} 
        \begin{subtable}{\linewidth}
            \centering
            \caption{The ablation study of different loss settings.}
            \label{tab:ablation_subb}
            \begin{tabular}{c|c|cccc|c}
            \toprule
            \multirow{2}{*}{Method} & \multirow{2}{*}{Exp. Index} & \multicolumn{4}{|c|}{Loss Settings} & \multirow{2}{*}{Performance}  \\
            &  & $\mathcal{L}_{pose}$ & $\mathcal{L}_{image}$ & $\mathcal{L}_{rvs}$  & $\mathcal{L}_{posemap}$ & \\
            \midrule
            \multirow{5}{*}{PMNet} 
            & 1 & \checkmark & \checkmark & \checkmark &            & 1.03~/~2.94 \\
            & 2 & \checkmark &            & \checkmark & \checkmark & 0.75~/~3.04 \\
            & 3 & \checkmark & \checkmark &            & \checkmark & 0.94~/~2.34 \\
            & 4 & \checkmark & \checkmark & \checkmark & \checkmark & \hl{\textbf{0.68~/~1.97}} \\
            \midrule
            \multirow{2}{*}{PMNet$_{ud}$}      
            & 5 &            & \checkmark &            &            & 0.39~/~0.63 \\
            & 6 &            & \checkmark &            & \checkmark & \hl{\textbf{0.31~/~0.55}} \\
            \bottomrule
            \end{tabular}
        \end{subtable}
    \end{minipage}
}
\end{table}

\subsection{Ablation Studies}
\label{subsec:ablation}

\textbf{Design choices of PoseMap.} 
Thanks to the neural volumetric pose feature, aka PoseMap, our method achieves more accurate estimation results than previous methods. 
The underlying strength of PoseMap lies in its ability to establish stable geometric correspondences between the image and the camera pose across diverse scenes.
To demonstrate the impact of PoseMap, we break down the pose features into three configurations in Table~\ref{tab:ablation_suba}. The baseline corresponds to the default settings of DFNet~\cite{chen2022dfnet} without any pose features supervision. $L_{nerfmap}$ refers to using the feature embedding extracted from the last MLP layer of pre-trained NeRF-H by directly rendering a feature map (termed as NeRFMap) for APRNet supervision. 
The comparatively weak results indicate that the original NeRF features do not inherently encapsulate camera pose information.
By incorporating a pose branch and a decoder, without modifying the original NeRF training protocol, PoseMap is able to learn the implicit features from the ground truth pose. 
With the assistance of PoseMap and the $L_{posemap}$ loss.

\begin{figure*}[t]
    \centering
    \includegraphics[width=\linewidth]{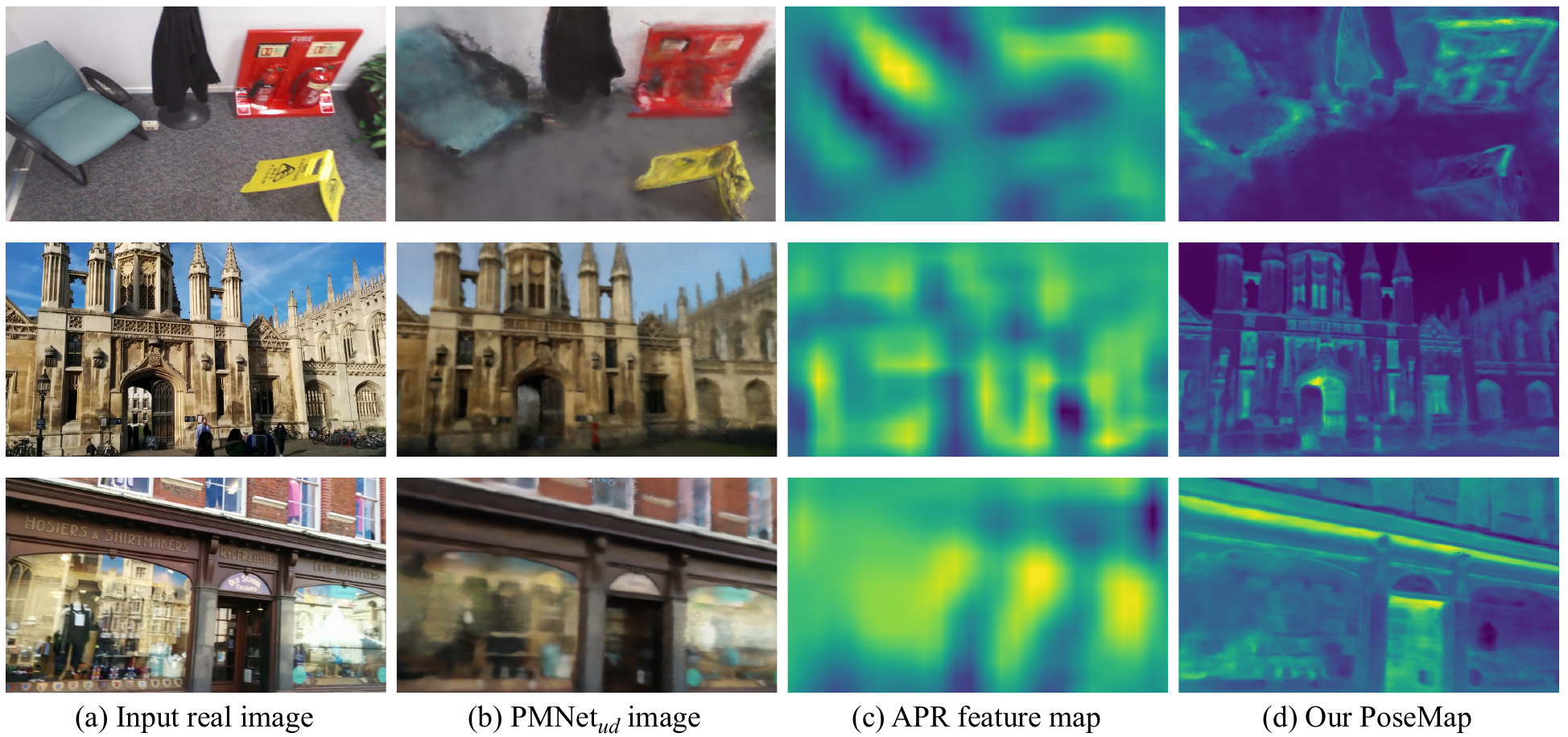}
    \caption{\textbf{Visualization of localization results.} From left to right, we show the input real image (left), the rendered image of the pose estimated by PMNet$_{ud}$ (2nd column), APR feature map (3rd column), and our PoseMap (right). The dimensionality reduction via PCA is utilized to visualize the PoseMap with pseudo color.}
    \label{fig:pose_vis}
\end{figure*}

\textbf{Effectiveness of the loss of PoseMap.} 
We conduct ablation studies on each loss component in Exp. 1, 2 and 3 to evaluate the individual contributions. $L_{image}$ acts as an effective feature adapter, bridging the domain gap between real and synthetic image features from our APRNet. It can yield a performance gain of 9.33\% in Exp.2 and 4. The RVS loss $L_{rvs}$, designed for novel view augmentation using a pretrained NeRF, shows an improvement of around 27.66\% in Exp.3 and 4. $L_{posemap}$ directs APRNet to learn implicit pose features, enhancing performance by 33.98\% in Exp.1 and 4 of PMNet and 20.51\% in Exp. 5 and 6. of PMNet$_{ud}$. The largest performance drop without $L_{posemap}$ also indicates the effectiveness of PoseMap. Additionally, for PMNet$_{ud}$, Exp.6 and 7 in Table~\ref{tab:ablation_subb} also highlight the significance of $L_{posemap}$ for feature alignment during the online refinement stage without labels.

\textbf{Visualization of PoseMap.} 
In Fig.~\ref{fig:pose_vis}, we visualize APR image features and PoseMap features. The APR image features are extracted from the output before the fully-connected layer of VGG\-16 network. Compared with APR image features, the proposed PoseMap features capture implicit features of the camera pose by aggregating global attributes from samples of light rays, resulting in local features with clearer geometric information than those from 2D-CNN backbones, which is useful on camera localization tasks as proven structure-based methods.

\begin{figure}[t]
    \centering
    \subfloat[The translation errors.]{
        \includegraphics[width=0.3\linewidth]{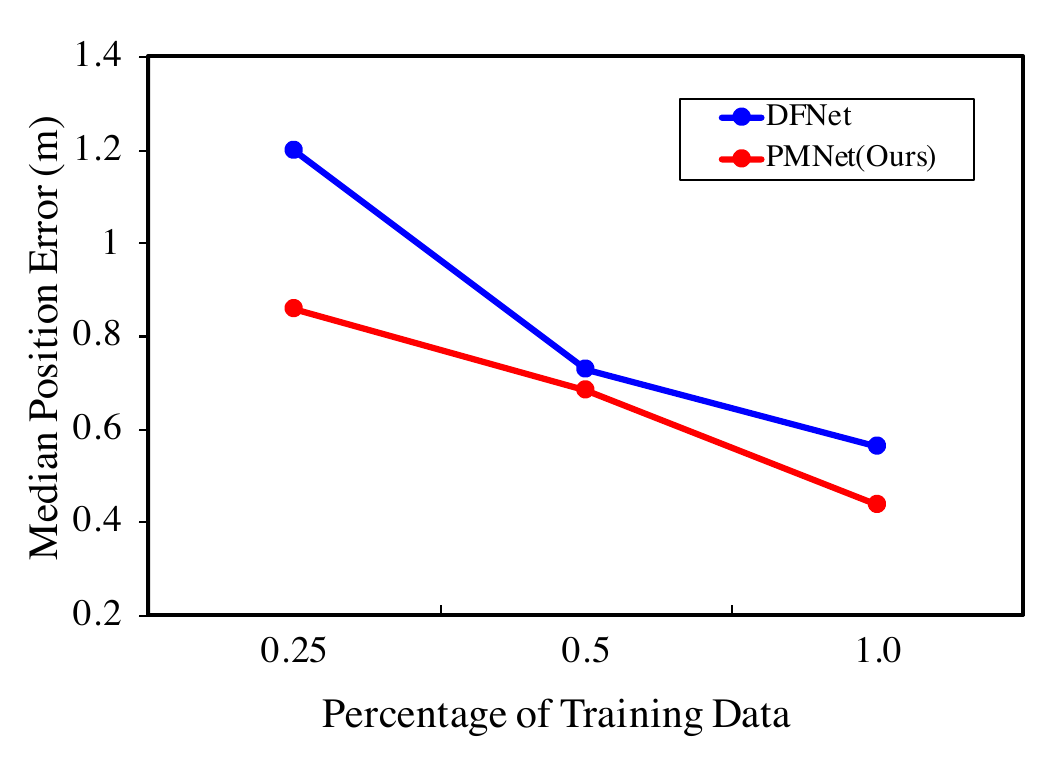}
    }
    \subfloat[The rotation errors.]{
        \includegraphics[width=0.3\linewidth]{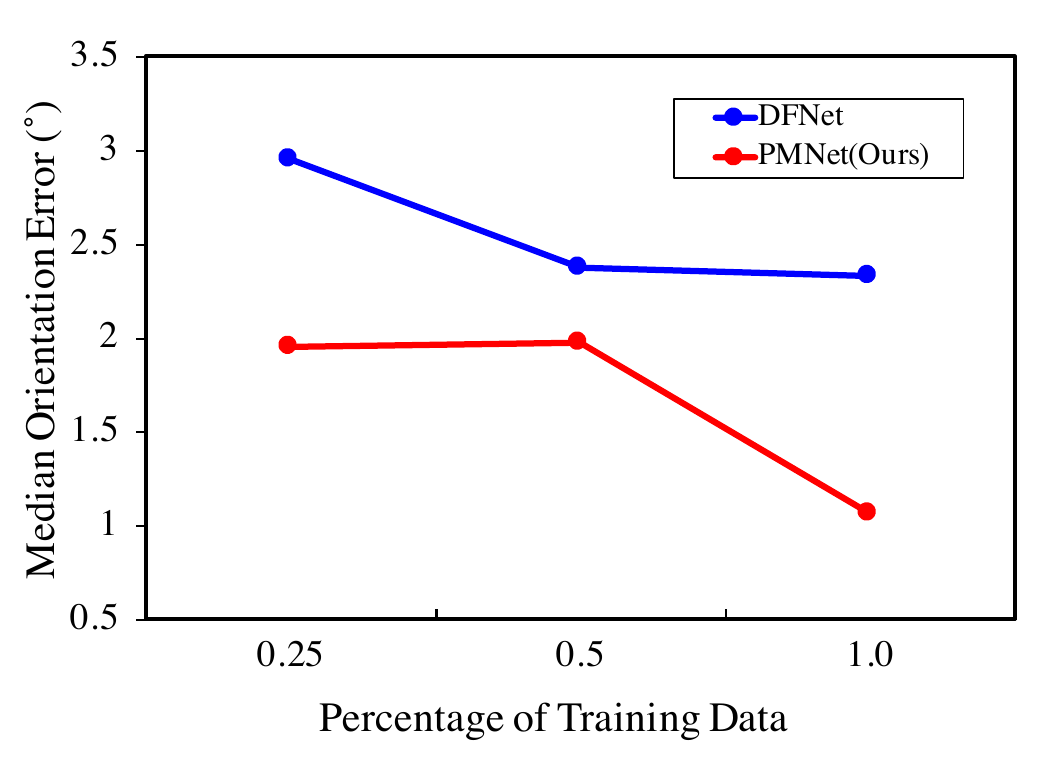}
    }
    \caption{\textbf{The quantitative results of training data size.} }
    \label{fig:datasize}
\end{figure}

\subsection{Discussions}
\label{subsec:exp_discussion}
\textbf{Robustness of training size.}
Since the effectiveness of camera localization correlates with the scale of the observed images~\cite{sattler2019understanding,moreau2022lens}, we perform an ablation experiment on the quantity of training data. We train the whole pipeline with 100\%, 50\%, and 25\% of the training set in Kings of Cambridge Landmarks dataset, the evaluations are shown in Fig.~\ref{fig:datasize}. As the data size decreases, the original DFNet\cite{chen2022dfnet} exhibits a rapid decrease in accuracy, indicating its high sensitivity to the quantity of input data. By comparison, our results are more robust to the size of training data as well as better performance.

\textbf{Discussion with related methods.}
As we state in Sec.~\ref{subsec:related_apr}, some post-processing refinement methods, \textit{e.g.}, CROSSFIRE~\cite{Moreau2023CROSSFIRECR} or NeFeS~\cite{Chen2023RefinementFA}, employ iterative pose refinement on reliable camera pose priors with implicit features matching or RANSAC-based PnP. Although it's unfair to directly compare our method with them, we are surprised to see that our method shows competitive performance in Table~\ref{table:cambridge} without any post-processing procedures, which also indicates the superiority of our PoseMap design.

\textbf{Limitations}. 
Similar to other learning-based methods, PMNet also shares limitations with NeRF and APRNet. Firstly, the accuracy of pose estimation is influenced by the quality of synthetic images, emphasizing the need for robust NeRF models to enhance results. Secondly, current APR-based approaches have yet to fully leverage the geometric structures present in images. Integrating more explicit structure cues, such as 2D lines and 3D depths, could be beneficial. Finally, at the inference stage, the adoption of hierarchical optimization techniques could refine the final results.
\section{Conclusion}
\label{sec:conclusions}
In this paper, we introduce PoseMap, a novel neural volumetric pose feature designed to enhance camera localization. This feature captures the implicit information of an image and its corresponding camera pose within a neural volumetric field and can be rendered by augmenting the original NeRF with a dedicated pose branch. We have developed a new APR architecture that incorporates a pose features extraction module for the task of camera localization. This architecture allows for online refinement with unlabelled data through self-supervision. Experiments show that our method achieves an average performance improvement of 14.28\% and 20.51\%, surpassing existing APR methods and establishing a new benchmark of accuracy. However, like other learning-based camera localization techniques, our method still falls short of the accuracy provided by structure-based methods. A potential avenue for future research is to integrate more structure-based features into our pipeline, as the demonstrated benefits of pose features are quite promising.
\section*{Acknowledgements}
This work was supported by the National Natural Science Foundation of China (62072422).


%
%
\bibliographystyle{splncs04}
\bibliography{main}

\clearpage
\setcounter{page}{1}
\section{Supplementary}
In the supplementary materials, we provide further insights of PoseMap designs in Sec.~\ref{subsec:sup_posemap}, implementation details in Sec.~\ref{subsec:sup_imple}, extended discussions in Sec.~\ref{subsec:sup_discuess},  and visualizations in Sec.~\ref{subsec:sup_visual}.

\subsection{Technical Details of PoseMap}
\label{subsec:sup_posemap}
Our motivation is to improve the performance of camera localization with a novel feature map called PoseMap.
Considering that NeRF itself takes the camera pose as input and conducts a volumetric rendering process. We surmise that it can help to identify an appropriate implicit feature representation for the camera pose.

We build our PoseMap based on the NeRF-W~\cite{martinbrualla2020nerfw} and NeRF-H~\cite{chen2022dfnet}. These two methods are designed for synthesizing novel views of complex scenes using only unstructured collections of in-the-wild photographs with different foreground occlusions and exposure levels. 
NeRF-W models the scene as a combination of both shared and image-specific elements, which allows for the unsupervised separation of scene's content into static and transient components. In our exploration of pose features, we recognize that the camera pose information is more linked with the static background. Thus, we design our pose branch to depend on these static elements. Within this branch, the pose embedding is then thoroughly distilled from the ground truth camera pose followed by a series of MLPs. To be specific, $\mathcal{F}_{image}$ comes from three CNN layers at $(\frac{H}{2},\frac{W}{2},128)$. $\mathcal{F}_{posemap}$ is rendered from the pose branch of NeRF-P at $(\frac{H}{16},\frac{W}{16},256)$. Both are pixel-wise features.

\begin{figure*}[t]
    \centering
    \includegraphics[width=\linewidth]{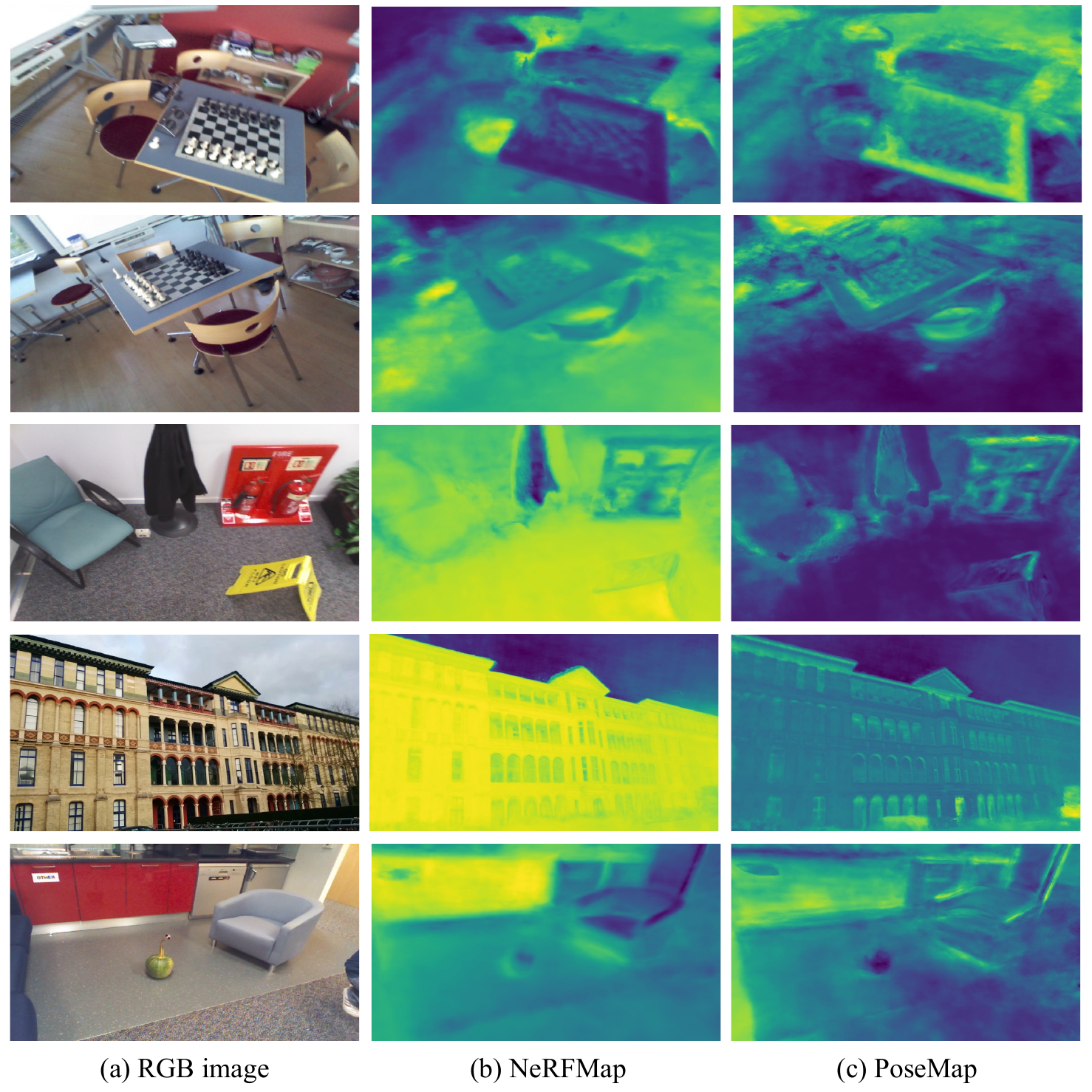}
    \caption{\textbf{Visualization comparison between NeRFMap and PoseMap.} From left to right, we show the RGB image (left), the NeRFMap (the middle column), and our PoseMap (right).}
    \label{fig:nerfmap_supl}
\end{figure*}

In order to explore the implicit information of PoseMap, We compare it with NeRFMap which is directly extracted from the original NeRF. As rendered and visualized in Fig.~\ref{fig:nerfmap_supl}, the salient region of PoseMap is more related to the geometrical structures such as lines and boundaries, which provides the obvious low-level evidence for camera poses estimation.

\subsection{Implementation Details}
\label{subsec:sup_imple}

\subsubsection{Datasets}
\label{subsubsec:dataset}
The numbers of frames of each scene in 7-scenes~\cite{shotton2013scene} and Cambridge Landmarks~\cite{kendall2015posenet} dataset are listed in Table~\ref{tab:dataset}. For a fair comparison with other methods, we follow the sampling strategy in DFNet to use subsampled training data with a spacing window $d=2$ in Cambridge Landmarks, $d=5$ in 7-scenes for scenes containing $\leq$ 2000 frames and $d=10$ otherwise. The ground-truth pose annotated in 7-scenes and Cambridge Landmarks are obtained from RGB-D SLAM method named dSLAM~\cite{6162880} and COLMAP~\cite{schoenberger2016sfm},  

\begin{table}[!]
    \begin{center}
    \caption{\textbf{The metadata of 7-scenes~\cite{shotton2013scene} and Cambridge Landmarks~\cite{kendall2015posenet}}}
    \label{tab:dataset}
    \scalebox{1.0}{
    \begin{tabular}{c|ccccccc}
    \toprule
    Scene &    Chess  &   Fire  &   Heads   &  Office  &  Pumpkin & Kitchen  &  Stairs   \\  
    \midrule
    train  &   4000  &  2000    &   1000    &   6000   &   4000   &   7000   &  2000   \\       
    test  &   2000  &  2000  &   1000    &   4000   &   2000   &   5000   &  1000       \\  
    \midrule
    Scene &   Kings   & Hospital &   Shop   & Church    \\
    \midrule
    train &   1220    &  895     &  231     &  1487   \\       
    test  &   343     &  182     &  103     &   530       \\ 
    \bottomrule
    \end{tabular}}
    \end{center}
\end{table}

\begin{table}[h]
    \begin{center}
    \caption{\textbf{The training time and memory cost of different stages.}}
    \scalebox{1.0}{
    \begin{tabular}{c|c|c|c}
    \toprule
    Stages &  NeRF-P  &  PMNet (APR / NeRF-P) & PMNet$_{ud}$ (APR / NeRF-P) \\ 
    \midrule
    Duration(hrs) & 13.0 & 18.1 (7.5 / 10.6) & 27.9 (11.1 / 16.8)  \\
    Memory(GB) & $\sim$5.0 & $\sim$12.8   & $\sim$18.4 \\
    \bottomrule
    \end{tabular}}
    \end{center}
\end{table}

\subsubsection{Training time and memory cost}
We report the training time and memory cost of \textit{Kings} from Cambridge Landmarks. The overall training takes about 2.4 days on a single NVIDIA RTX 4090 GPU, including NeRF, PMNet, PMNet$_{ud}$. To be specific, APR training time makes up around $31.5\%$(18.6hrs), the training and rendering of NeRF account for nearly $68.5\%$(40.4hrs). Notably, NeRF-based methods (DFNet, LENS, etc.) have extended training duration due to extensive computational burden of NeRF itself. We plan to opt for more efficient rendering schemes in the future. However, inference time is not affected. Besides, excluding NeRF's rendering time, the APR training time for PMNet (7.5 hrs) slightly exceeds that of DFNet (6.0 hrs) due to the computational demands of PoseMap. We firmly believe the notable improvement in performance, as in quantitative and qualitative results, outweighs the slight increase in computational resources.

\subsubsection{Inference}
\label{subsubsec:runtime}
For the inference time, our method runs at 1.82ms (about 550Hz) for a $240\times427$ image and 6.27ms (about 160Hz) for a $480\times854$ image. 
As for training duration, since we have incorporated NeRF into our online training pipeline, the time required for training is largely dominated by the volumetric rendering process.
All our experiments are conducted on an NVIDIA GeForce RTX 4090 GPU with PyTorch.

\subsection{Discussions}
\label{subsec:sup_discuess}

\subsubsection{Comparing with refinement methods}
Unlike APR-based methods that directly estimate the camera pose, the iterative refinement-based methods require prior estimation of the camera pose and continuously refine it based on the consistency of pixel features. Generally, these methods can achieve higher accuracy in pose estimation. In the experiments, as shown in Table 3 of the paper, our approach, PMNet$_{ud}$, achieves comparable accuracy of pose estimation to refinement-based methods. This suggests that, due to the design of our PoseMap and the domain consistency in implicit pose features, our algorithm can effectively enhance the precision of APR-based methodologies. In certain scenarios, it even outperforms refinement-based methods, demonstrating greater practical utility. 

\subsubsection{Comparing with SfM$/$SLAM methods}

As illustrated in Sec.~\ref{subsubsec:dataset}, most datasets of camera localization are annotated by SfM or SLAM. SfM aims to recover 3D geometry and camera parameters from image sets, which is done classically by computing local image features, finding matches across images and then estimating and verifying epipolar geometry using bundle adjustment. SLAM is used for localizing and mapping the surroundings from a video stream by matching feature descriptors in adjacent frames. As noted in Table~\ref{table:summary_camloc} of the paper, the main differences between SfM-based methods and APR-based methods are two-fold. First, SfM-based methods are the stage of in-sample building with a set of in-sample images with unknown poses, which is built to obtain the poses of in-sample images. However, APR-based methods are in the stage of out-of-sample testing, aiming to use in-sample images to train a practical network for out-of-sample camera pose regression. Second, as the ground truth is obtained with COLMAP (over all frames of each scene) in the Cambridge Landmarks dataset, we believe the accuracy of all APR related methods would be limited by this upper bound. Therefore, we claim that it's unfair to evaluate the performance comparisons with these methods.

\subsection{Visualization}
\label{subsec:sup_visual}

\begin{figure*}[h]
    \centering
    \includegraphics[width=\linewidth]{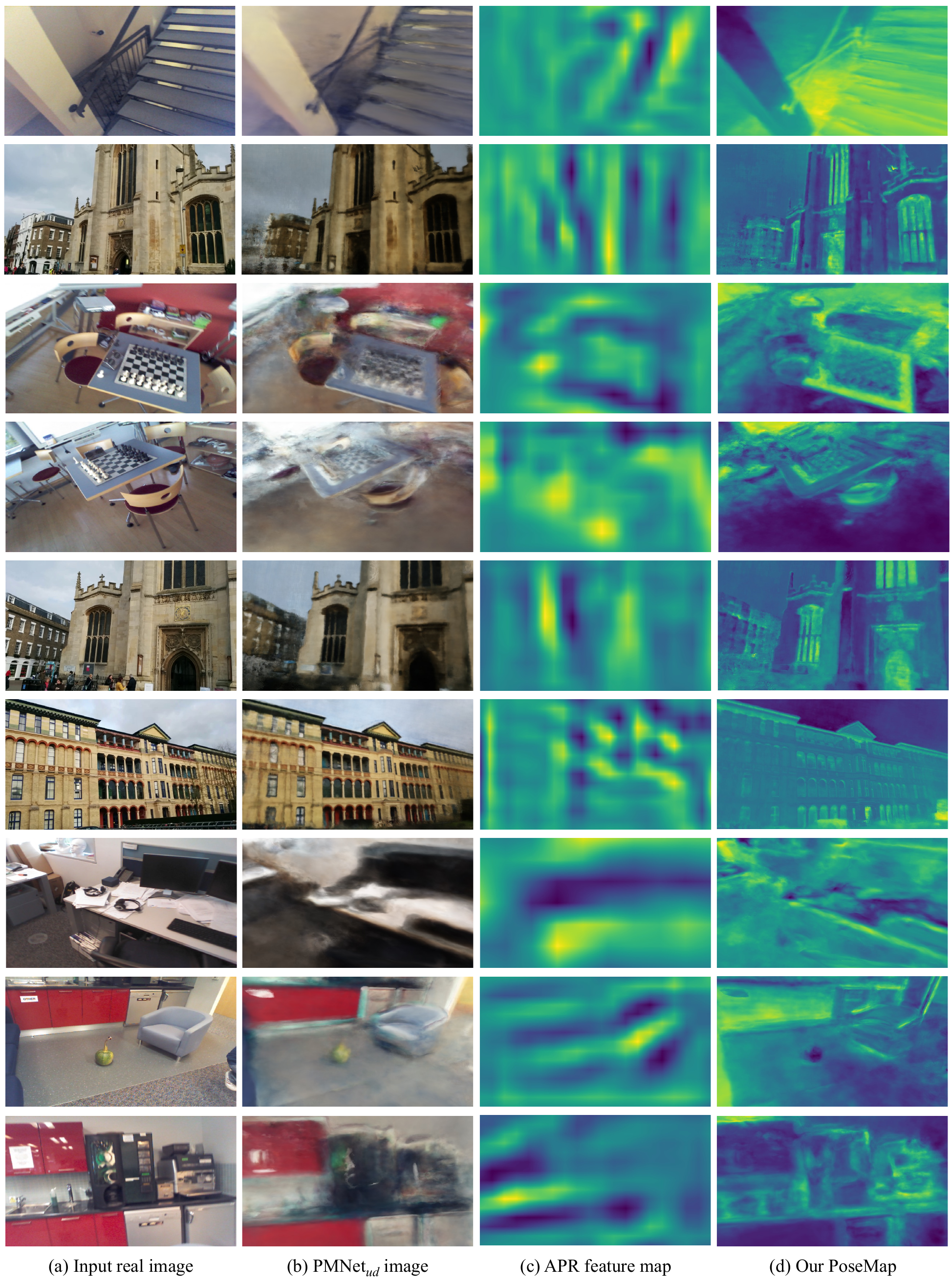}
    \caption{\textbf{Visualization of more localization results.} We show (a) the input real image, (b) the rendered image using the pose estimated by PMNet$_{ud}$, (c) the APR feature map and (d) our PoseMap. Note that, The dimensionality reduction via PCA is used to visualize the feature map.}
    \label{fig:pose_vis_supl}
\end{figure*}

Here, we present more visualization samples of PoseMap in Fig.~\ref{fig:pose_vis_supl}. As we can see, APR feature map is a CNN-based feature map pre-trained on ImageNet datasets. The salient region is more related to the semantics and appearance. However, our PoseMap is an MLP-based feature map created through volumetric rendering, which captures geometric details more effectively for improved pose estimation. Additionally, the supervision of APR feature map faces the challenge of a domain gap between real images and rendered images, an issue that doesn't apply to the PoseMap.

\end{document}